\documentclass{IEEEtran}
\usepackage[utf8]{inputenc}
\usepackage{microtype}
\usepackage{graphicx}
\usepackage{epstopdf}
\usepackage{amsmath,amssymb,booktabs,tabularx}
\usepackage{bm}
\usepackage[hyphens]{url}
\usepackage{hyperref}
\hypersetup{
    colorlinks,
    linkcolor={red!75!black},
    citecolor={blue!75!black},
    urlcolor={blue!75!black}
}

\usepackage{cleveref}
\crefname{section}{Sec.}{Sections}
\crefname{figure}{Fig.}{Figure}
\crefname{table}{Tab.}{Table}
\crefname{equation}{Equ.}{Equation}
\usepackage{mathrsfs}
\usepackage{float}
\usepackage{multirow}
\usepackage{soul}
\usepackage[table]{xcolor}
\usepackage{comment}
\usepackage{subcaption}
\usepackage{multicol}
\usepackage{import}
\usepackage{siunitx}
\usepackage{mathtools}
\usepackage{xspace}
\usepackage{makecell} 

\newcommand{\ie}{i.\,e.\xspace}
\newcommand{\eg}{e.\,g.\xspace}
\newcommand{\cf}{\emph{cf}.\xspace}

\newcommand{\bx}{\bm{x}}
\newcommand{\by}{\bm{y}}
\newcommand{\bg}{\bm{g}}

\newcommand{\bW}{\bm{W}}
\newcommand{\bb}{\bm{b}}
\newcommand{\fx}{F_x}
\newcommand{\fg}{F_g}
\newcommand{\fint}{F_{\text{int}}}

\newcommand{\h}[1]{\textbf{#1}}

\title{Glacier Calving Front Segmentation Using Attention U-Net}

\author{\IEEEauthorblockN{
		Michael Holzmann\textsuperscript{1*}, 
		Amirabbas Davari\textsuperscript{1*}, 
		Thorsten Seehaus\textsuperscript{2}, 
		Matthias Braun\textsuperscript{2}, 
		Andreas Maier\textsuperscript{1},\newline
		Vincent Christlein\textsuperscript{1}\\
	}
	\textsuperscript{1} Department of Computer Science, Friendrich-Alexander
	Universität Erlangen-Nürnberg, Germany\\ 
	\textsuperscript{2} Department of Geography \& Geosciences, Friedrich-Alexander Universität Erlangen-Nürnberg, Germany\\ 
	* Michael Holzmann and Amirabbas Davari contributed equally to this work.
}
\begin{document}
\maketitle

\begin{abstract}
An essential climate variable to determine the tidewater glacier status is the location of the calving front position and the separation of seasonal variability from long-term trends. Previous studies have proposed deep learning-based methods to semi-automatically delineate the calving fronts of tidewater glaciers. They used U-Net to segment the ice and non-ice regions and extracted the calving fronts in a post-processing step.
In this work, we show a method to segment the glacier calving fronts from SAR images in an end-to-end fashion using Attention U-Net. The main objective is to investigate the attention mechanism
in this application. Adding attention modules to the state-of-the-art U-Net network 
lets us analyze the learning process by extracting its attention maps. We use these maps as a tool to search for proper hyperparameters and loss functions in order to generate higher qualitative results. Our proposed attention U-Net performs comparably to the standard U-Net while providing additional insight into those regions on which the network learned to focus more. In the best case, the attention U-Net achieves a \SI{1.5}{\percent} better Dice score compared to the canonical U-Net with a glacier front line prediction certainty of up to 237.12 meters.

\end{abstract}


\section{Introduction}\label{sec:introduction}
The Antarctic Peninsula is a hotspot of global warming. It has experienced a substantial temperature increase in the 20th century~\cite{absence}, strongly affecting its cryosphere. In 1995, the Prince-Gustav-Channel and Larsen-A ice shelves (floating extension of the glaciers) disintegrated~\cite{climatictrend}. The loss of ice shelves led to a reduction of the bracing forces on the tributary glaciers, which subsequently reacted by accelerating ice discharge and further retreat of the calving fronts~\cite{dynamicresponse,icedynamics,retreat}. The position of the calving front is an important variable to measure the glacier state. Its fluctuations can provide information about imbalances and a recession can destabilize the flow regime of an entire glacier system. Therefore monitoring the calving front positions is of high importance. 

Typically multi-spectral and synthetic aperture radar (SAR) remote-sensing imagery is used to map the position of the glacier calving fronts~\cite{remotesensing}. Manual delineation of the calving front position is applied in most studies, which is subjective, laborious, tedious and expensive~\cite{glacieraccuracy}. However, different (semi-)automatic routines using edge detection or image classification techniques were also developed. Baumhoer et al.~\cite{remotesensing} provide a detailed review on calving front detection approaches. Particularly in polar regions, the sea surface next to the calving front is often covered by the ice melange (sea ice and icebergs), making both the manual and automatic calving front detection challenging. 

Deep Convolutional Neural Networks (CNNs) have shown impressive performance in various image segmentation tasks. 
The first study using deep CNNs for calving front mapping on remote sensing imagery was conducted by Mohajerani et~al.~\cite{margins}. They applied the U-Net CNN architecture on Landsat imagery at glaciers in Greenland. Zhang et~al.~\cite{enze19} and Baumhoer et~al.~\cite{glacierextraction} applied similar processing pipeline on SAR imagery from either the TerraSAR-X or Sentinel-1 mission.
Today, there are many deep learning-based approaches and network architectures for image segmentation~\cite{DeepLearnSeg}. One of the most successful ones is the U-Net architecture
~\cite{Unet}. 
In 2018, Schlemper et~al.~\cite{Sononet} used soft attention gates in Sononet, which are able to localize salient regions in the feature maps. They introduced these gates to help the network preserve more local information and provide guided object localization in the forward pass. These gates use low-level features and global information gathered by the high-level feature maps in order to calculate an attention coefficient for each pixel. This approach was further enhanced by Oktay et~al.~\cite{AttUNet}. They added additive soft attention gates in the skip connections of the U-Net architecture. These gates learn the attention maps automatically without adding too much computational power. They are modular and can easily be implemented in the U-Net architecture. We use these gates together with the same U-Net model used in Zhang et~al.'s work~\cite{enze19} to detect the front positions on multi-mission SAR imagery. 
Predicting thin fronts directly leads to a severe class-imbalance problem. 
In this work, we extract the attention maps for each layer to see how well the network learns and where its attention is. Furthermore, we study the effect of distance weighted loss functions to alleviate this problem. 


\section{Methodology}\label{sec:methodology}
~\\\noindent\textbf{Attention U-Net.}\noindent
\begin{figure*}[t]
	\centering
	\includegraphics[width=0.8\linewidth]{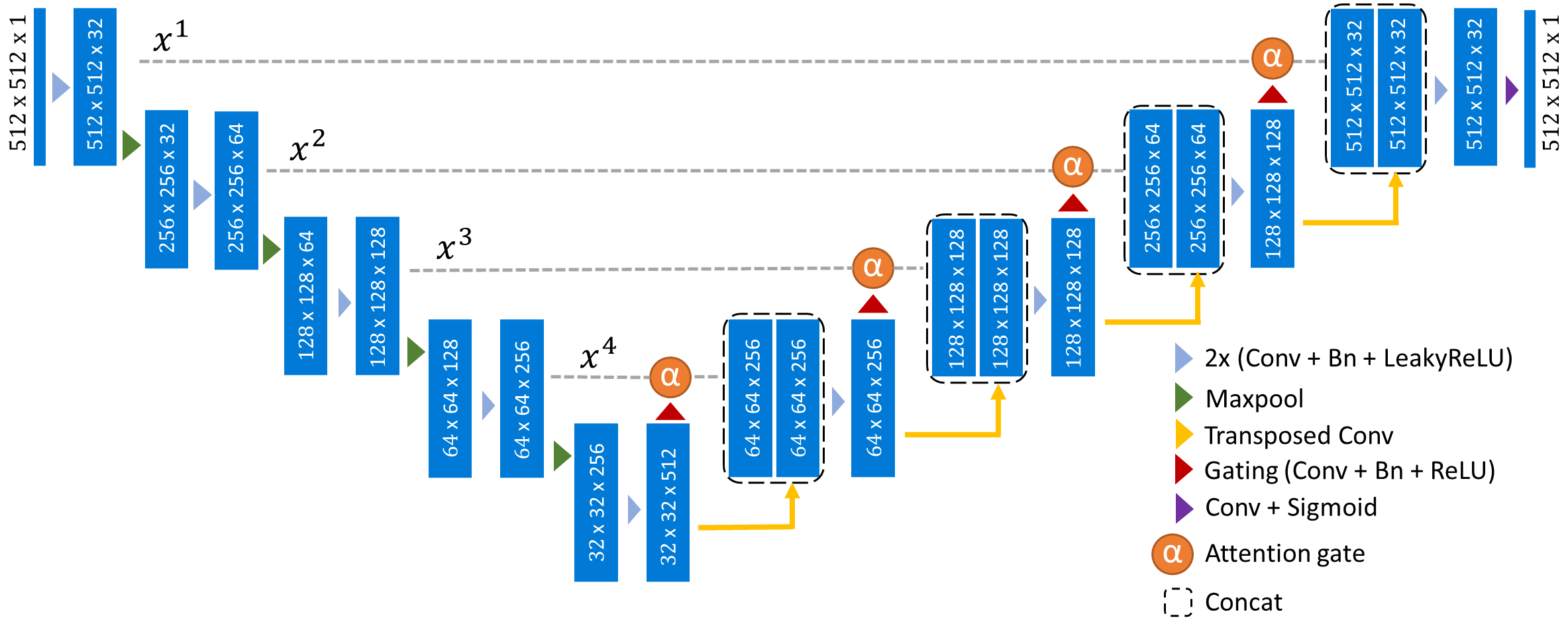}
	\caption{The architecture of our proposed Attention U-Net. }
	\label{fig:attunet}
\end{figure*}
%
Attention gates are modular building blocks, which can be placed into the skip
connections of a U-Net architecture~\cite{AttUNet}. We use a five-layered U-Net
model by Zhang et al.~\cite{enze19} as our base model, depicted in
\cref{fig:attunet}. 
Each layer has two sequential blocks consisting of one two-dimensional convolution with a kernel size of $5 \times 5$ pixels and one batch normalization layer followed by a Leaky ReLU activation function. We use Max-Pooling to encode and shrink the feature maps on each layer and use transposed convolution in the upscaling path. 

We add attention gates to the same architecture and train and test it with the same data and hyperparameters.
Features (activation maps) $\bx^l$ 
are copied to the attention gates together with the gating signal $\bg^l$ taken
from the upscaling path. The number of feature maps $\fx$ of $\bx^l$ and $\fg$ of $\bg^l$ are reduced to $\fint$ intermediate layers by means of $1 \times 1$
convolution. Due to the coarser scale of the gating signal, the spatial
dimensions of $\bx^l$ are halved using strided convolution. To combine low-level
and high-level features maps, the intermediate layers are added and passed to a
ReLU activation function $\sigma_1$.
In order to obtain the attention coefficients, these layers are
combined through $\bm{\psi} \in \mathbb{R}^{\fint \times 1}$ (realized by a $1 \times
1$ convolution) and normalized by the Sigmoid
activation function $\sigma_2$, \ie for each pixel $i$ the attention coefficient
for layer $l$ follow as: 
%
\begin{equation}\label{eq:att_coef}
	\alpha^l_i =\sigma_2 (\bm{\psi}^T (\sigma_1 (\bW^T_x \bx^l_i + \bW^T_g \bg^l_i + \bb_g) + b_\psi) \enspace,
\end{equation}
where $\bW_x \in \mathbb{R}^{\fx \times \fint}$ and $\bW_g \in \mathbb{R}^{\fg \times \fint}$ are pixel-wise linear combinations of $\fx$ and $\fg$ feature maps and transformations to $\fint$ intermediate layers. The bias terms are denoted by $\bb_g \in \mathbb{R}^{\fint}$ and $b_\psi \in\mathbb{R}$. In the end, they are upscaled by a bilinear interpolation to the original spatial dimensions of the incoming feature maps. The attention coefficients $\alpha^l$ represent context-relevant regions. We apply the soft attention mechanism by pixel-wise multiplication of the attention maps and every feature map $x^l$, \ie: 
%
	$\hat{x}^l_{i,c} = x^l_{i, c} \cdot \alpha^l_i$, 
%
where $c$ describes the feature map index. 
In this way, noisy and irrelevant regions in the feature maps get filtered out by the attention gate before going through the merging process. 
During training, those attention maps that localize the regions of interest are learned automatically.

In this work, we use Zhang et~al.'s U-Net architecture~\cite{enze19} as the baseline model in our semantic segmentation pipeline and add the attention gates to it, as depicted in \cref{fig:attunet}. 
The attention gates narrow down the region of interest by automatically learning the glacier front line positions. Furthermore, the easily extractable attention maps demonstrate the learning quality evolution of the network.
%

\begin{figure}[t]
	\centering
	\begin{subfigure}[b]{0.20\linewidth}		
		\centering
		\centerline{\includegraphics[width=.87\linewidth]{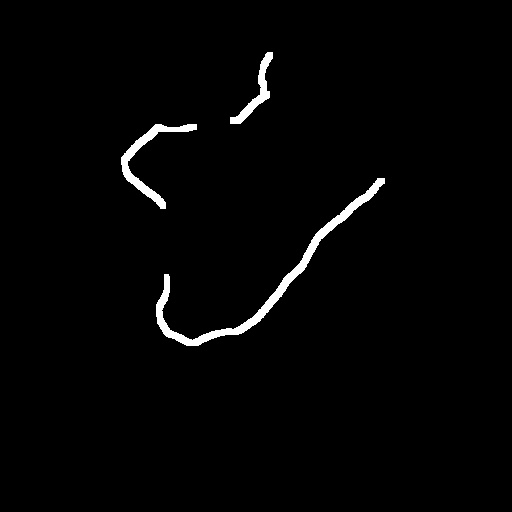}}
		\caption{}\label{fig:dw_a}
	\end{subfigure}
	\begin{subfigure}[b]{0.20\linewidth}		
		\centering
		\centerline{\includegraphics[width=1\linewidth]{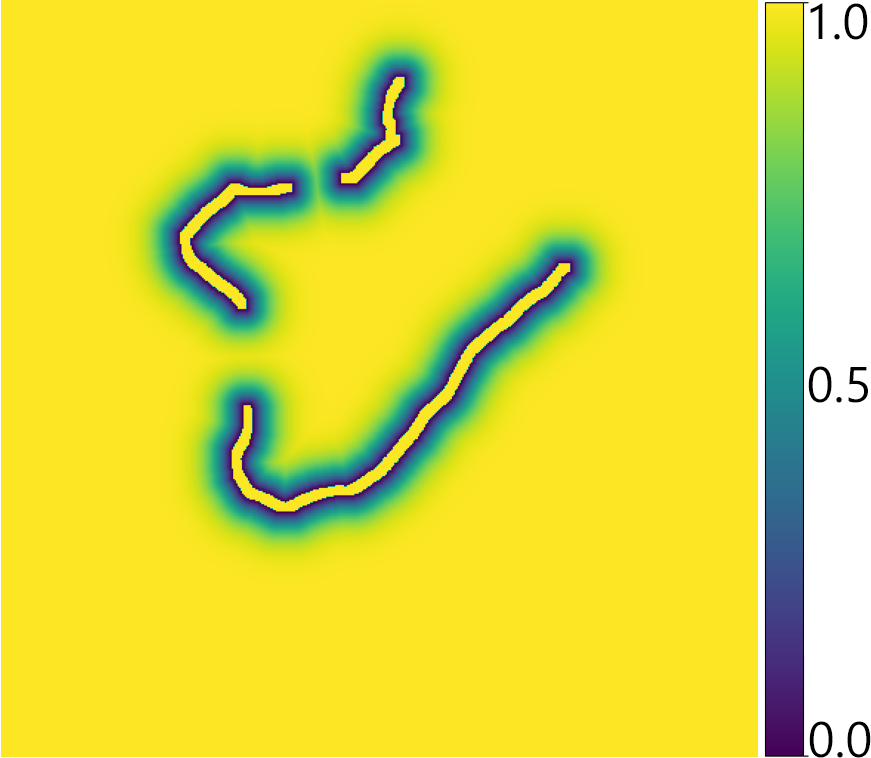}}
		\caption{}\label{fig:dw_b}
	\end{subfigure}
	\begin{subfigure}[b]{0.20\linewidth}		
		\centering
		\centerline{\includegraphics[width=.87\linewidth]{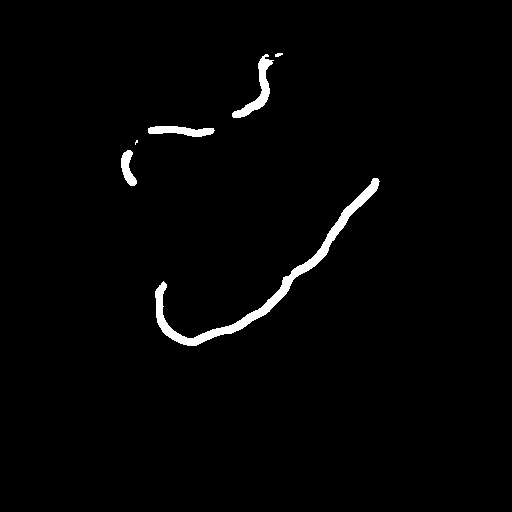}}
		\caption{}\label{fig:dw_c}
	\end{subfigure}
	\begin{subfigure}[b]{0.20\linewidth}		
		\centering
		\centerline{\includegraphics[width=1\linewidth]{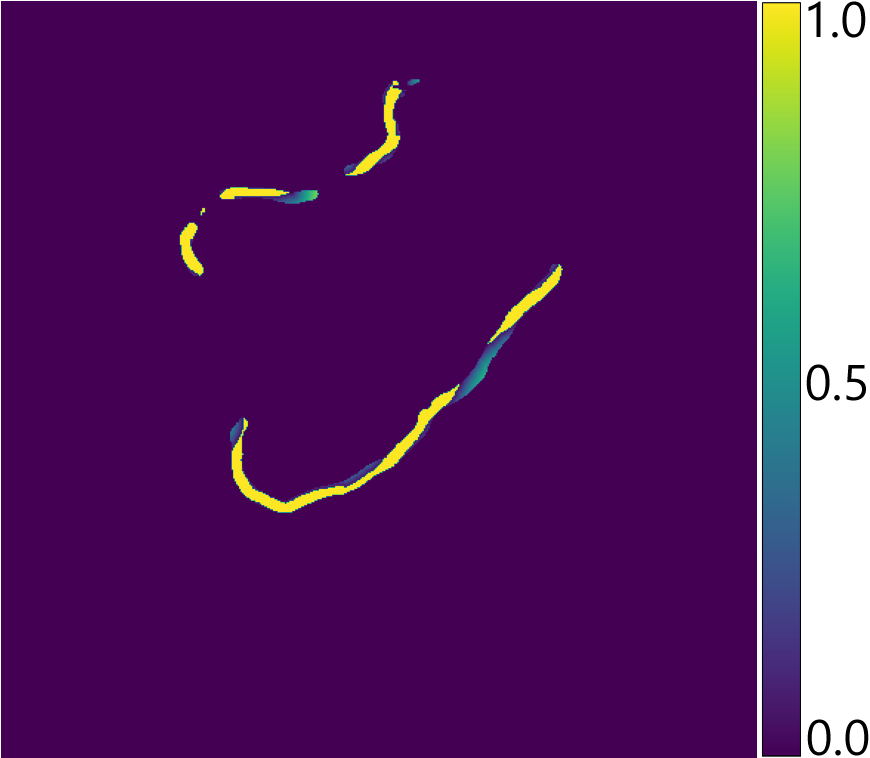}}
		\caption{}\label{fig:dw_d}
	\end{subfigure}
	\caption{\subref{fig:dw_a} ground truth, \subref{fig:dw_b} weight map $\widetilde{\bW}_8$, \subref{fig:dw_c} prediction, and \subref{fig:dw_d} weighted prediction $\hat{\by}_8$.}
	\label{fig:distance_weighting}
\end{figure}

~\\\noindent\textbf{Distance-Weighted Loss Function.}\label{sec:dwloss}
\noindent
One big challenge in segmenting the front line directly is that a slight offset in the prediction heavily increases the loss. 
Adhikari et~al.~\cite{foresttrail} proposed to tolerate the network's
misprediction in a controlled close distance of the ground truth front line.
This is formulated and realized by calculating a weight map $\bW_w$:
\begin{equation}\label{eq:Ww}
	\bW_w = \sigma\left(\frac{\text{EDT}(\by)}{w}\right) \enspace,
\end{equation}
%
which depends on the Euclidean distance transformation $\text{EDT}$ of the
ground truth map $\by$. Errors near the front line get weighted smaller than errors
further away. The weights are normalized by a Sigmoid function $\sigma$. With the parameter $w$, we can control the distance weighting. Bigger $w$
results in tolerance towards the mispredictions further away from the actual
front line. In this study, we explore weights $w =
\{4,8,16\}$. 
%
%
The sigmoid output is zero-centered and added to the ground truth mask in order to omit weighting pixels within the glacier front, \ie:
%
	$\widetilde{\bW}_w = 2(\bW_w - 0.5) + \by$. 
%
The result of this transformation is shown in \cref{fig:dw_b}. This distance map contains the weighting coefficient for each pixel of the prediction $\hat{\by}$. We pixel-wise multiply the prediction and the distance-weighted map. The result is a distance-weighted prediction $\hat{\by}_w$, 
\cf \cref{fig:dw_d} for an example.
%
%
We use this modified prediction for the calculation of the BCE loss and non-binary Dice metric.
%
%

\section{Evaluation}\label{sec:experimental_setup}
\subsection{Dataset}\label{sec:dataset}
The SAR dataset is composed of imagery from the satellite missions ERS-1/2, Envisat, RadarSAT-1, ALOS, TerraSAR-X (TSX) and TanDEM-X (TDX). It contains $244$ images showing the glacier systems of Sjögren-Inlet (SI) and Dinsmoore-Bombardier-Edgworth (DBE) located in the Antarctic Pensula. Together, they cover the time period of 1995-2014. We use multilooking to reduce speckle in the image and take the refined ASTER digital elevation model~\cite{cook2012new}  of the Antarctic Peninsula for geocoding and orthorectification. 
%
%
We split this dataset into $144$ samples for training, $50$ for validation and $50$ for testing. As for the expert data annotation, we exploited the front positions used by Seehaus et al.~\cite{icedynamics, dynamicresponse}. 
%
\begin{figure}[t]
	\centering
	\mbox{%
	\includegraphics[width=\linewidth]{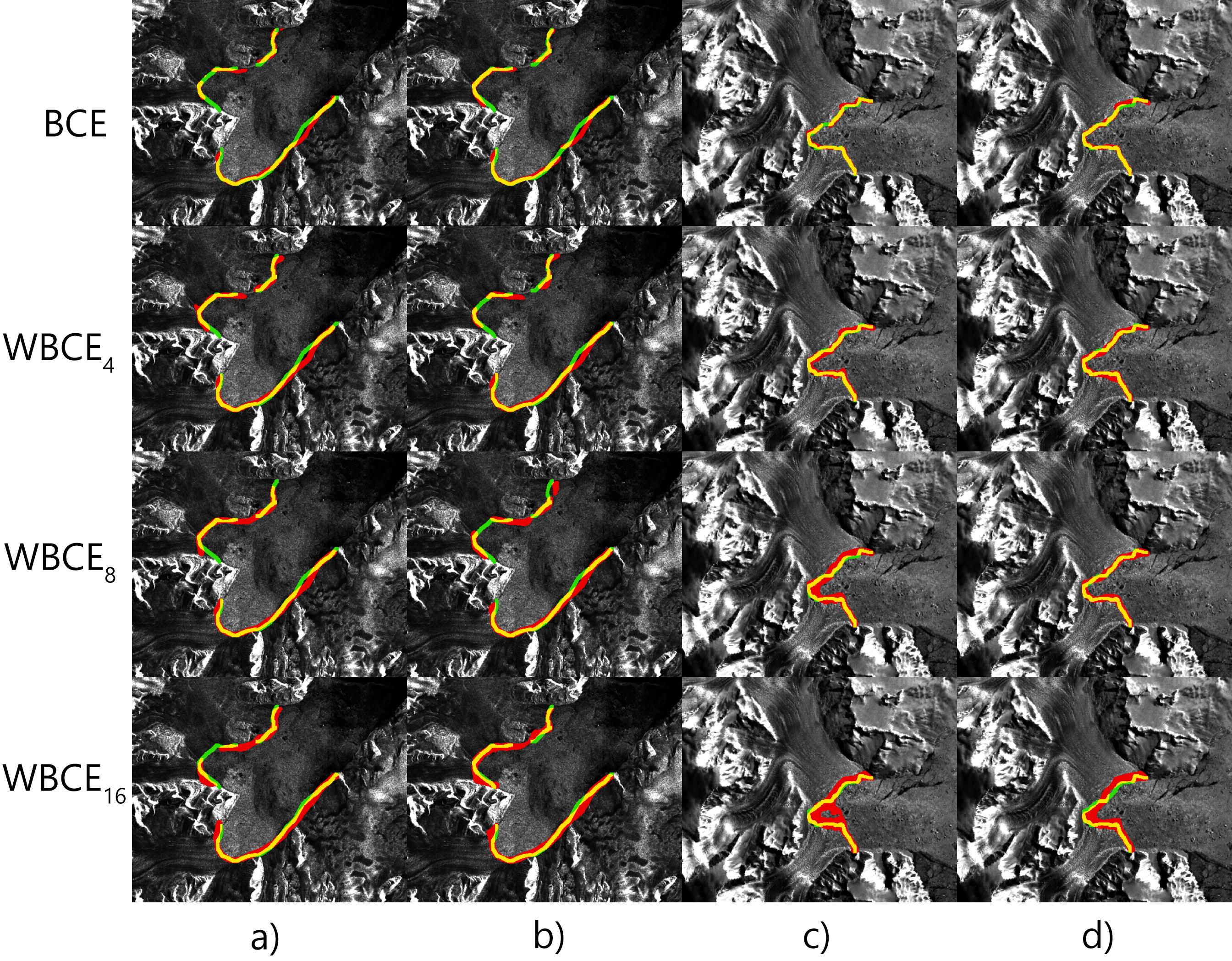}
	\phantomsubcaption\label{fig:p_a}	\phantomsubcaption\label{fig:p_b}	\phantomsubcaption\label{fig:p_c}	\phantomsubcaption\label{fig:p_d}
	}%
	\caption{Predictions of both networks trained on different weights for two samples of the test set. \subref{fig:p_a} and \subref{fig:p_c} are predicted by the U-Net approach~\cite{enze19}, and \subref{fig:p_b} and \subref{fig:p_d} show the results produced by our Attention U-Net; green: ground truth, red: prediction, yellow: intersection of the ground truth and the prediction.}
	\label{fig:predictions}
\end{figure}
\subsection{Evaluation Protocol}
We applied median filtering for noise reduction. Since the images in the dataset have different spatial resolutions, we used an adaptive kernel size, such that it covers a region of \SI{2500}{m^2}. Next, we applied zero-padding on the images to form squares and then resized the images to $512 \times 512$ pixels using bilinear interpolation. In order to alleviate the severe class-imbalance problem, we used morphological dilation to thicken the glacier front lines in the ground truth images such that they became $6$ pixels wide after downscaling. This reduces the class-imbalance severity from approximately 2000:1 to 100:1. Additionally, we augmented our training set using vertical flips and rotated versions ($90^\circ$, $180^\circ$ and $270^\circ$).

For training the networks, we used Adaptive Moment Estimation (ADAM) optimizer
with a batch size of $5$ and a cyclic learning rate with minimum and maximum boundaries of $10^{-5}$ and $10^{-3}$, respectively. 
For the Leaky ReLU activation functions, we set the slope to $0.1$. 
As for the loss function, we used the normal Binary Cross-Entropy (BCE) in one variant
and the distance-weighted BCE (WBCE) in the other variant. We monitored the Dice coefficient and its modified version for early stopping with a patience of $20$ epochs. 


\subsection{Qualitative Results} \label{sec:qualitativ}
\Cref{fig:predictions} illustrates the predictions of two glacier fronts using U-Net and Attention U-Net, trained on a distance-weighted BCE loss with different weights. 
It shows that our proposed Attention U-Net outperforms U-net for big distance weights, especially for $w=16$, as shown in \cref{fig:predictions}. 
Note, increasing $w$ results in thicker predicted lines and consequently higher uncertainty. 
%
%
Using the normal BCE loss often results in disconnected predicted lines. While WBCE mitigates this problem, this can sometimes be a disadvantage, \eg when mountains near the front lines  cause gaps on the ground truth line due to layover and shadowing. \Cref{fig:p_a} and \cref{fig:p_b} show that networks trained on weight parameter greater than four fail to localize small gaps and simply connect the predicted fronts. 

\begin{figure}[t]
	\centering
	\includegraphics[width=\linewidth]{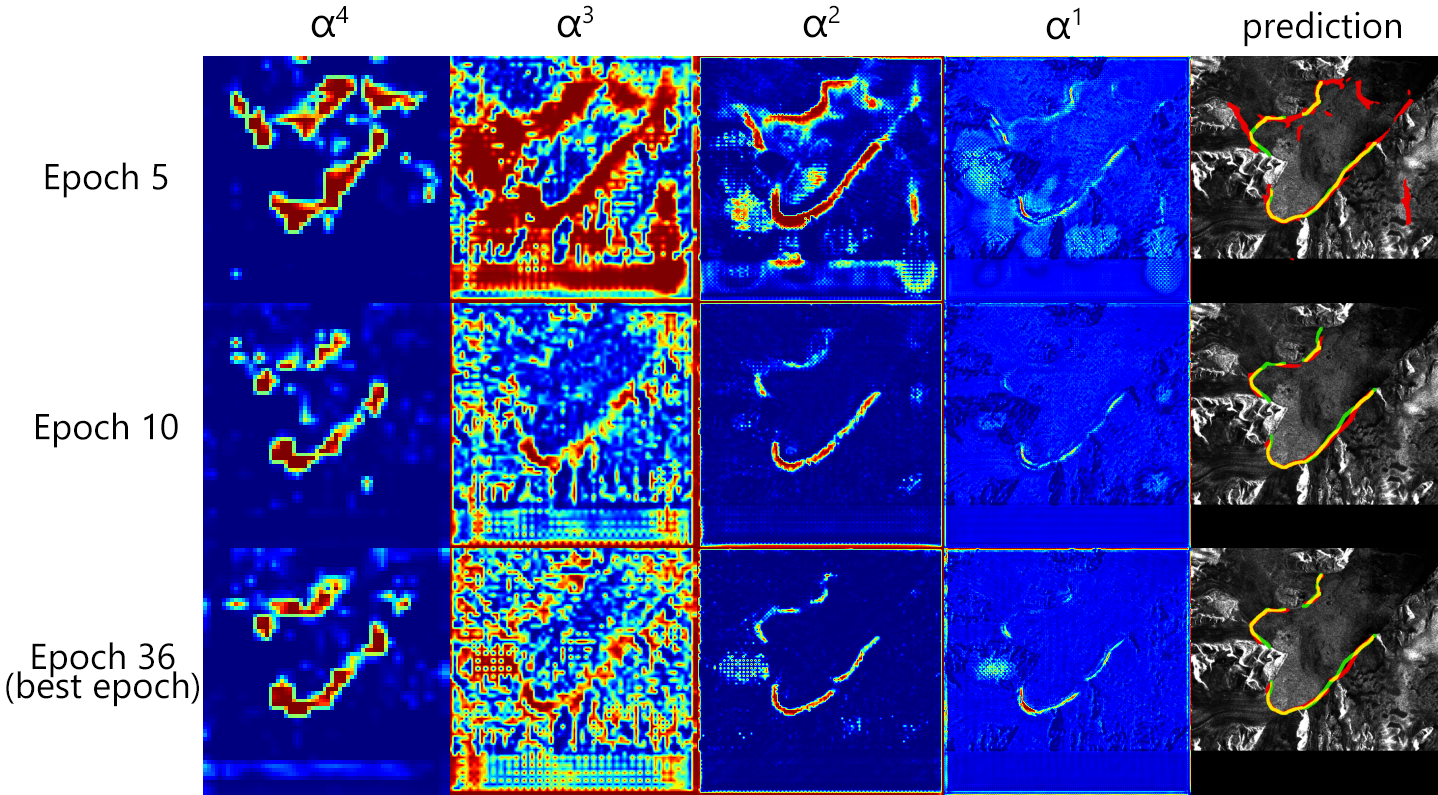}
	\caption{Attention maps $\alpha^l$ for models saved at epoch 5, 10, and 36.}
	\label{fig:attention_2007_1_1}
\end{figure}

We can analyze the learning process of Attention U-Net by means of the attention maps $\alpha^l$, extracted from layer $l$. 
\Cref{fig:attention_2007_1_1} shows the evolution of the attention maps (resized to the same scale). 
We observe that the attention maps converge quickly within the first epochs to the regions of interest. In general, attention maps $\alpha^4$ and $\alpha^3$ look more noisy than $\alpha^2$ and $\alpha^1$. However, this shows that the attention gates can learn salient regions automatically, even in the bottom layers of a U-Net. 
Besides the glacier front, there is also a weak highlighting on the zero-padded border and other regions like mountains and ice melange located near the front. This may help the network to recognize specific glacier systems. 

\begin{figure}[t]
	\centering
	\mbox{%
	\includegraphics[width=.6\linewidth]{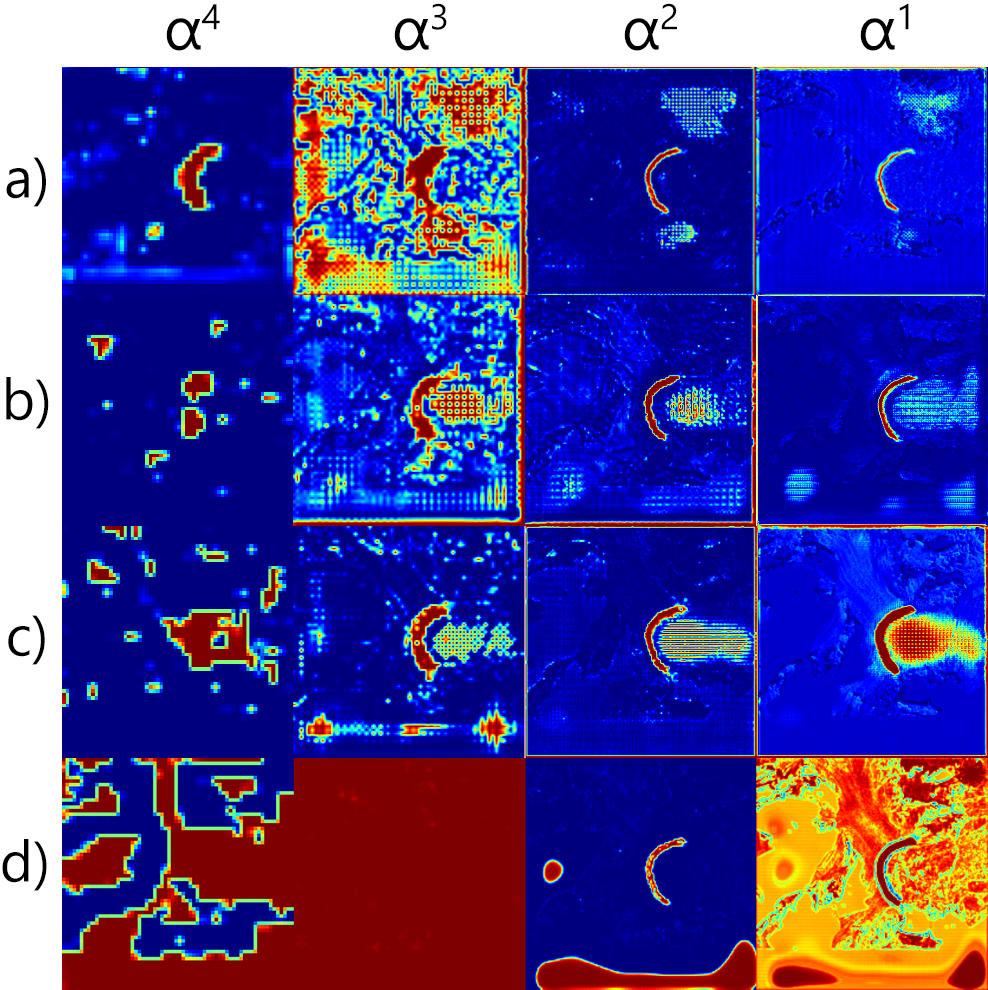}
	\phantomsubcaption\label{fig:a_a}	\phantomsubcaption\label{fig:a_b}	\phantomsubcaption\label{fig:a_c}	\phantomsubcaption\label{fig:a_d}
}%
\caption{Attention maps for models trained on \subref{fig:a_a} BCE, \subref{fig:a_b} $\mathrm{WBCE}_4$, \subref{fig:a_c} $\mathrm{WBCE}_8$, and \subref{fig:a_d} $\mathrm{WBCE}_{16}$.}
	\label{fig:attention_maps_weights}
\end{figure}

\Cref{fig:attention_maps_weights} shows the effect of different distance weights on the attention maps. Training on a distance-weighted loss function with a small weight reduces noisy highlights in the attention maps, especially in layer $3$. When increasing the weight too much, the attention gates cannot localize the regions of interest in the low-level feature maps, as shown in \cref{fig:a_d}. 
Choosing a weight between $w=4$ and $w=8$ lead to the best qualitative results in our test set.

\subsection{Quantitative Results}\label{sec:quantitativ}
\begin{table}[t]
	\centering
	\caption{Quantitative results of the U-Net, with and without the attention
		gates.%
}
	\small
	\begin{tabular}{@{ }c@{ }@{ }c@{ }@{ }c@{ }@{ }c@{ }@{ }c@{ }@{ }c@{ }@{ }c@{ }}
		\toprule
		Model & Loss & Dice & $\mathrm{WDice}_4$ & $\mathrm{WDice}_8$&
		$\mathrm{WDice}_{16}$ & IoU \\ 
		\midrule
		U-Net & BCE &\h{73.9} & \h{83.0} & \h{86.0} & 88.3 & \h{60.6} \\
			Att. U-Net & BCE &73.8 & 82.9 & 85.9 & \h{88.7} & 60.3 \\
		\midrule
		U-Net & $\mathrm{WBCE}_4$ &\h{72.9} & \h{83.6} & \h{87.1} & \h{90.1} & \h{58.9} \\
		Att. U-Net & $\mathrm{WBCE}_4$ &72.6 & \h{83.6} & 87.0 & 89.9 & 58.5 \\
		\midrule
		U-Net & $\mathrm{WBCE}_8$ &70.1 & 81.1 & 84.6 & 87.6 & 55.5 \\
		Att. U-Net & $\mathrm{WBCE}_8$ &\h{70.8} & \h{82.5} & \h{86.1} & \h{89.2} & \h{56.1} \\
		\midrule
		U-Net & $\mathrm{WBCE}_{16}$ &\h{67.0} & \h{78.9} & \h{82.6} & 85.9 & \h{51.9} \\    
		Att. U-Net & $\mathrm{WBCE}_{16}$ &65.2 & 77.8 & 82.3 & \h{86.1} & 49.8 \\ 
		\bottomrule
	\end{tabular}
	\label{tab:quantitativ}
\end{table}

\Cref{tab:quantitativ} presents quantitative results for our test set. 
Overall, the results show that Attention U-Net performs similar to the base U-Net with the advantage of better interpretability. In case of a distance weight of 8 (WBCE$_8$), Attention U-Net performs significantly better than the base model. Distance-weighted Dice scores with bigger weights are generally better since errors near the front are weighted smaller.
\begin{table}[t]
	\centering
	\caption{Average thickness of the predicted lines in pixels and the resulting certainty (perpendicular to the estimated glacier front position) based on an average spatial resolution of \SI{44.74}{m}.}
	\small
	\begin{tabular}{cccccc}
		\toprule
		$w$ & 0 & 4 & 8&16 \\ 
		\midrule
		thickness [px] & 6.5 & 8.4 & 10.6 &  11.1  \\ 
	  certainty [m]  & 145.40 & 187.90 & 237.12 & 248.30 \\ 
		\bottomrule
	\end{tabular}
	\label{tab:uncertainty}
\end{table}
Assuming a small amount of misprediction, we can estimate the average thickness of the lines by the class ratio of front to the background. \Cref{tab:uncertainty} shows that training on BCE with different weights leads to front predictions with different thicknesses and certainties.
%
%
With $w=4$, we obtain a Dice score of \SI{83.6}{\percent} with a tolerance of $187.90$ meters. It is possible to reduce this uncertainty by using a smaller dilation kernel, smaller distance weight, or a patch-wise line segmentation with a higher pixel resolution.


\section{Conclusion}\label{sec:conclusion}
In summary, this study shows that it is possible to delineate calving glacier fronts by means of a trained U-Net in a single forward pass without additional post-processing. The addition of attention gates revealed insights into the learning behavior of the network. Within a few epochs, the network learns to localize important regions automatically and shows where its attention is. Our results show that the attention mechanism itself improves our predictions for big distance weights. Furthermore, it can be used as an analysis tool to find proper loss functions and hyperparameters. Consequently, we conclude that training on a distance weighted version of Binary Cross-Entropy with a small weight parameter leads to optimal qualitative and quantitative results.


\bibliographystyle{IEEEbib}
\bibliography{References}

\end{document}